\pdfoutput=1
\documentclass[conference]{IEEEtran}
\usepackage{times}
\usepackage[numbers]{natbib}
\usepackage{multicol}
\usepackage[bookmarks=true]{hyperref}
\hypersetup{hidelinks}
\usepackage{booktabs}
\usepackage{graphicx}
\usepackage{amsmath}
\usepackage{tikz}
\usepackage{colortbl}
\usepackage{pgfplots}
\pgfplotsset{compat=1.18}
\usepgfplotslibrary{groupplots}
\usetikzlibrary{positioning,arrows.meta,calc,shapes.geometric}
\ifdefined\BUILD\else\def\BUILD{arxiv}\fi
\newif\ifanon
\def\rssbuild{rss}
\ifx\BUILD\rssbuild \anontrue \else \anonfalse \fi
\newif\ifappendix \def\arxivbuild{arxiv}\ifx\BUILD\arxivbuild \appendixtrue \else \appendixfalse \fi
\newcommand{\paperid}{[add Paper-ID]}
\newif\ifbrand\brandtrue
\newcommand{\dch}{ElevateHead}\newcommand{\titleprefix}{ElevateHead: }

\ifanon
\hypersetup{pdftitle={\titleprefix Decide Once, Then Commit: When the Action Head Matters for Multimodal Imitation}, pdfauthor={}, pdfsubject={}, pdfkeywords={}, pdfcreator={}, pdfproducer={}}
\else
\hypersetup{pdftitle={\titleprefix Decide Once, Then Commit: When the Action Head Matters for Multimodal Imitation}, pdfauthor={Tsung-Lung Yang}, pdfsubject={Robot learning}, pdfkeywords={imitation learning, multimodality, action heads, commitment}}
\fi

\ifanon\newcommand{\later}{Stage~2}\newcommand{\Later}{Stage 2}\else\newcommand{\later}{a follow-up version of this paper}\newcommand{\Later}{Planned Real-Robot Test}\fi
\newcommand{\ci}[2]{\mbox{[#1,\,#2]}}

\begin{document}

\title{\titleprefix Decide Once, Then Commit:\\When the Action Head Matters for Multimodal Imitation}

\ifanon
\author{Author Names Omitted for Anonymous Review. Paper-ID \paperid}
\else
\author{\authorblockN{Tsung-Lung Yang}
\authorblockA{National University\\
Email: webureetaipei@gmail.com}}
\fi

\maketitle

\begin{abstract}
Robots that learn from people often see several valid ways to do the same task: pass an obstacle on the left or on the right, open the microwave or move the kettle first. A policy that re-decides at every step can waver between these ways mid-task. We ask when the \emph{action head} (the part of a policy that turns the state into actions) matters, and how it should make this choice. On a controlled task, heads that sample one way succeed far more often than a head that averages the ways when the demonstrations contain several ways and the state gives no hint which to take ($+0.72$ success with two ways); this trend was not reproduced on the other tasks we ran. We then present \dch{}, a head that \emph{decides once} at a learned decision point, \emph{commits} to that choice for the episode, and exposes the choice as an input a user can set. We designed it \emph{by diagnosis}: each version was frozen and tested on tasks it had not seen, and the next version changed only what a measured failure pointed to; predictions were registered before each run. The first four versions each fell below BeT on at least one held-out task. The fifth completed all four subtasks of the held-out Franka Kitchen task in 45.2\% of episodes against 29.3\% for BeT, the strongest baseline across our tasks ($+0.160$, 95\% CI \ci{+0.129}{+0.188}); when a mode was forced, the first completed subtask matched it in 95\% of successful episodes. Diagnosis also located where commitment breaks: varied start positions (repaired on the controlled task: commitment 0.22 to 0.81, success 0.74 to 0.91) and varied object placement (partly repaired: in an exploratory variant, commitment 0.03 to 0.59 on aligning with no detectable change in success). On held-out block stacking, version 4 was comparable to BeT and did not exceed it. A real-robot test with predictions fixed in advance is planned for \later.
\end{abstract}

\IEEEpeerreviewmaketitle

\section{Introduction}
\begin{figure}[t]
\centering
\begin{tikzpicture}[font=\scriptsize, x=1.02cm, y=1.35cm]
\foreach \dx/\ttl in {0/{(a) re-decide every step}, 2.85/{(b) decide once, then commit}, 5.7/{(c) user sets the mode}}{
  \begin{scope}[xshift=\dx cm]
  \fill[gray!40] (1.0,-0.2) rectangle (1.4,0.2);
  \fill (0,0) circle (1.5pt); \node[below, font=\tiny] at (0,-0.04) {start};
  \draw[thick] (2.4,0) circle (2.5pt); \node[below, font=\tiny] at (2.4,-0.06) {goal};
  \node[font=\scriptsize\bfseries, align=center, text width=2.7cm] at (1.2,0.85) {\ttl};
  \draw[gray!60, densely dotted] (0.6,0) .. controls (0.9,0) and (0.85,0.42) .. (1.2,0.42) .. controls (1.55,0.42) and (1.6,0) .. (2.3,0);
  \draw[gray!60, densely dotted] (0.6,0) .. controls (0.9,0) and (0.85,-0.42) .. (1.2,-0.42) .. controls (1.55,-0.42) and (1.6,0) .. (2.3,0);
  \end{scope}}
\draw[very thick, red!75!black, -{Latex[length=1.4mm]}] (0.05,0) -- (0.6,0) -- (0.72,0.22) -- (0.8,-0.16) -- (0.88,0.18) -- (0.92,-0.05);
\node[red!75!black, align=center] at (1.2,-0.66) {switches ways, stalls};
\begin{scope}[xshift=2.85cm]
\draw[very thick, blue!70!black, -{Latex[length=1.4mm]}] (0.05,0) -- (0.6,0) .. controls (0.9,0) and (0.85,0.42) .. (1.2,0.42) .. controls (1.55,0.42) and (1.6,0) .. (2.28,0);
\node[diamond, draw, fill=yellow!60, inner sep=1.4pt] at (0.6,0) {};
\node[blue!70!black, align=center] at (1.2,-0.66) {choice kept to the goal};
\end{scope}
\begin{scope}[xshift=5.7cm]
\draw[very thick, orange!85!black, -{Latex[length=1.4mm]}] (0.05,0) -- (0.6,0) .. controls (0.9,0) and (0.85,-0.42) .. (1.2,-0.42) .. controls (1.55,-0.42) and (1.6,0) .. (2.28,0);
\node[diamond, draw, fill=yellow!60, inner sep=1.4pt] at (0.6,0) {};
\node[orange!85!black, align=center] at (1.2,-0.66) {``go below'' is followed};
\end{scope}
\end{tikzpicture}
\caption{The idea in one picture. Dotted: the two demonstrated ways around the obstacle (gray). (a) A head that samples its choice again at every step can switch ways mid-route. (b) \dch{} chooses once, at a learned decision point (diamond), and keeps the choice. (c) Because the choice is an explicit input, a user can set it.}
\label{fig:teaser}
\end{figure}
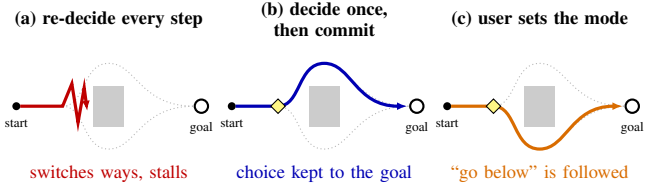

Imitation learning trains a robot policy from human demonstrations. When people solve a task in several ways, the data contain several correct actions for the same situation, and the \emph{action head}---the last layer of the policy---has to pick one. A head that averages the demonstrations (regression) can output an action that belongs to no way at all, such as driving straight into the obstacle between a left and a right detour. Heads that sample one way avoid the average---discrete selection as in BeT~\citep{shafiullah2022bet} and VQ-BeT~\citep{lee2024vqbet}, diffusion~\citep{chi2023diffusion}, flow matching~\citep{lipman2023flow}---but they usually sample again at every step and can switch ways mid-route (Fig.~\ref{fig:teaser}a). On common benchmarks the heads are hard to tell apart, and recent analyses disagree on why~\citep{mazza2026multimodality,pan2025muchado,zeng2026openloop}.

This paper makes three contributions:
\begin{enumerate}
\item \textbf{When the head matters.} On a controlled task, the head matters when the demonstrations contain several ways \emph{and} the state gives no hint which to take; the gap shrinks once such a hint exists (Section~\ref{sec:when}). The trend was not reproduced on the other tasks we ran, and we report that.
\item \textbf{A head that decides once and commits.} \dch{} learns \emph{where} demonstrations branch, chooses a way there only if the choice is genuinely open, keeps it for the episode, and exposes it as an input (Fig.~\ref{fig:teaser}b,c; Section~\ref{sec:method}).
\item \textbf{Design by diagnosis, with the full record.} Every version was frozen, tested on tasks it had not seen, and, when it fell short, diagnosed before anything was changed (Section~\ref{sec:design}). After four versions that fell below BeT somewhere, the fifth exceeded BeT on held-out Franka Kitchen ($p_4$ 0.452 vs.\ 0.293; Section~\ref{sec:results}). The diagnoses also yield lessons for other heads (Section~\ref{sec:lessons}).
\end{enumerate}
\begin{table}[t]
\centering
\caption{Main findings at a glance.}
\label{tab:glance}
\setlength{\tabcolsep}{2pt}\scriptsize
\begin{tabular}{@{}p{0.26\columnwidth}p{0.355\columnwidth}p{0.345\columnwidth}@{}}
\toprule
Question & Answer & Key evidence\\
\midrule
When does the head matter? (\S\ref{sec:when}) & With several ways and no hint in the state (controlled task); not reproduced on other tasks & gap $+0.72$ at two ways, $-0.04$ when the average path works\\
Does deciding once help on an unseen task? (\S\ref{sec:results}) & Yes on Kitchen; not shown on stacking & $p_4$ 0.452 vs.\ 0.293 for BeT, $+0.160$ \ci{+0.129}{+0.188}\\
Does the mode input set the behavior? & On Kitchen, mostly (descriptive) & first subtask matches the forced mode in 95.4\% of successes\\
Where does commitment break? & Varied starts (repaired on one task), few demonstrations & commitment 1.00 to 0.22; gate (\ref{eq:gate}) restores 0.81\\
Varied object placement (toward clutter)? & Partly repaired; success unchanged within uncertainty (exploratory) & commitment 0.03 to 0.59 (aligning), 0.32 to 0.64 (stacking)\\
Which baseline is strongest? (Table~\ref{tab:allheads}) & BeT (or tied) on every task, among eight heads & Kitchen $p_4$: BeT 0.293, flow 0.111, diffusion 0.000\\
\bottomrule
\end{tabular}
\end{table}
\textbf{Highlights.} (i) On held-out Kitchen, full-task success rises from 29.3\% (BeT) to 45.2\%, about 1.5$\times$; the gain over version~4 ($+0.222$, descriptive) comes from two inference rules, with no retraining. (ii) Forcing a mode sets the first subtask in 95.4\% of successful episodes---a control input BeT lacks. (iii) A diagnosed gate fix lifts commitment under varied starts from 0.22 to 0.81 and cuts route switches by 74\% (1.38 to 0.35 per episode). (iv) Under varied object placement, closer to cluttered scenes, commitment rises from 0.03 to 0.59 (aligning) and from 0.32 to 0.64 (stacking) in an exploratory variant. (v) A chunk costs 0.63--0.66\,ms, 2.6--4.0$\times$ faster than our flow and diffusion heads (Table~\ref{tab:allheads}). Table~\ref{tab:glance} summarizes the findings. Every registered prediction was committed to a time-stamped log before its run; post-hoc analyses are labelled as such, and every held-out result against BeT is shown (Fig.~\ref{fig:record}).

\section{Key Terms and Related Work}\label{sec:terms}
\textbf{Terms.} A policy maps the state $c_t$ (positions of arm and objects) to an \emph{action chunk}, the next $H{=}10$ actions, which are executed before the policy is queried again. A \emph{mode} (a ``way'') is a family of similar demonstrations, such as all left detours. A \emph{decision point} is a state where the demonstrations split into several modes. \emph{Committing} means keeping one mode until the episode ends. A task is \emph{held-out} for a version if none of its data was used to design that version (always within this study); the other tasks are \emph{development} tasks. A test is \emph{registered} if its prediction and decision rule were logged before the run.

\textbf{Related work.} Implicit~\citep{florence2021ibc}, diffusion~\citep{chi2023diffusion}, flow~\citep{lipman2023flow,black2025pi0} and one-step~\citep{geng2025meanflow} heads sample continuous actions, also inside chunked~\citep{zhao2023act} and vision-language-action policies~\citep{kim2024openvla}; BeT and VQ-BeT select a discrete code at every step. \citet{mazza2026multimodality} relate generative behavior cloning to multimodality and average feasibility; \citet{pan2025muchado} attribute its gains to iteration and stochasticity; \citet{zeng2026openloop} explain long open-loop execution by non-Markovian experts. MARS~\citep{jia2026mars} doses multimodality by a learned dispersion signal, GeoAAC~\citep{chen2026geoaac} adapts how long a chunk runs, and bidirectional decoding~\citep{liu2025bid} keeps consecutive chunks consistent at test time. Latent plans~\citep{lynch2019play} and options~\citep{sutton1999options} also hold a choice over time, and inference-time steering~\citep{wang2025itps} sets a diffusion policy's behavior from user input. \dch{} differs in deciding \emph{where} to choose from mode counts in the data and in holding the choice for the episode.

\section{When Does the Action Head Matter?}\label{sec:when}
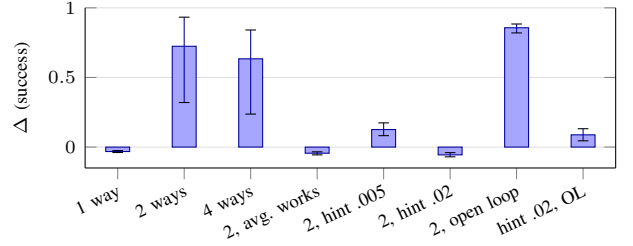
\begin{figure}[t]
\centering
\begin{tikzpicture}
\begin{axis}[width=8.6cm, height=3.7cm, ybar, bar width=9pt, ymin=-0.15, ymax=1.0,
  symbolic x coords={k1,k2,k4,k2F,j005,j02,k2OL,j02OL},
  xtick=data, xticklabels={{1 way},{2 ways},{4 ways},{2, avg.\ works},{2, hint .005},{2, hint .02},{2, open loop},{hint .02, OL}},
  tick label style={font=\scriptsize}, x tick label style={rotate=25, anchor=north east, inner sep=1pt}, ylabel={$\Delta$ (success)}, label style={font=\scriptsize},
  ymajorgrids, grid style={gray!25}, enlarge x limits=0.07, axis x line*=bottom]
\addplot[fill=blue!35, draw=blue!60!black, error bars/.cd, y dir=both, y explicit]
 coordinates {(k1,-0.032) +- (0,0.007) (k2,0.724) -= (0,0.404) += (0,0.209) (k4,0.634) -= (0,0.397) += (0,0.207)
 (k2F,-0.044) -= (0,0.012) += (0,0.010) (j005,0.126) -= (0,0.044) += (0,0.048) (j02,-0.056) -= (0,0.014) += (0,0.016)
 (k2OL,0.857) -= (0,0.037) += (0,0.027) (j02OL,0.088) -= (0,0.043) += (0,0.044)};
\end{axis}
\end{tikzpicture}
\caption{When the head matters (controlled task, 5 seeds, 95\% CI). $\Delta$: success of the five sampling heads minus that of an averaging (L1) head. ``avg.\ works'': two ways whose average path also reaches the goal; ``hint'': random start offset that tells which way to take; open loop (OL): no re-planning. The gap appears with several ways and disappears when averaging works or a hint exists.}
\label{fig:delta}
\end{figure}

\textbf{Setup.} A controlled 2-D task: reach a goal past an obstacle along 1, 2 or 4 demonstrated ways (400 demonstrations each). We compare L1 regression, MIP~\citep{pan2025muchado} and five heads that sample: autoregressive, diffusion, flow matching, discrete diffusion and MeanFlow~\citep{geng2025meanflow}. Let $\Delta$ be the mean success of the five sampling heads minus that of L1.

\textbf{Finding (Fig.~\ref{fig:delta}).} With one way the gap is near zero ($-0.032$; registered equivalence test inconclusive). With two ways averaging fails: $\Delta{=}{+0.724}$ \ci{+0.320}{+0.933}, and the registered slope over $\log_2$ of the number of ways is $+0.333$ \ci{+0.134}{+0.437} (supported). The gap vanishes when the average of the two ways is itself a working path ($-0.044$; registered contrast $+0.768$ \ci{+0.363}{+0.981}), and it shrinks when a small start offset hints at the way ($+0.126$ and $-0.056$). It is larger without re-planning (open loop; $+0.143$ \ci{+0.097}{+0.189} at offset 0.02). Re-planning more often can also rescue averaging heads (Section~\ref{sec:lessons}, finding~2).

\textbf{Limit.} Predicting $\Delta$ for new tasks from their measured multimodality missed on robomimic~\citep{mandlekar2021robomimic} (six of six datasets), D3IL avoiding and sorting (cross-task slope $-0.375$, descriptive). On the controlled task the gap appears above one way; on the non-synthetic tasks we ran, the trend was not reproduced, and a test on a real arm is planned for \later.

\textbf{What this means for head design.} When the state gives no hint, a head must \emph{select} a way; once selected, the way should be \emph{kept}, because sampling again invites switching. This is the principle behind \dch{}.

\section{\dch{}: Decide Once, Then Commit}\label{sec:method}
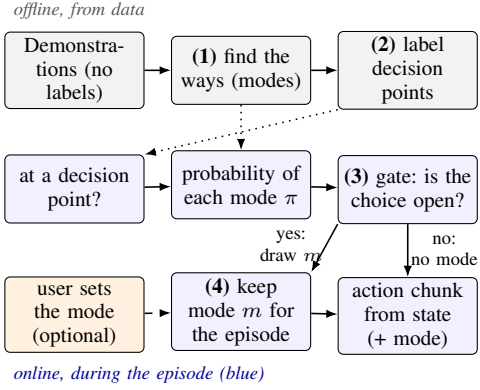
\begin{figure}[t]
\centering
\begin{tikzpicture}[font=\footnotesize,
  box/.style={draw, rounded corners=2pt, align=center, inner sep=2pt, minimum height=9mm, text width=17mm},
  train/.style={box, fill=gray!10}, infer/.style={box, fill=blue!6},
  arr/.style={-{Latex[length=1.5mm]}, semithick}, lab/.style={font=\scriptsize, align=center, inner sep=1pt}]
\node[train] (demo) {Demonstra\-tions (no labels)};
\node[train, right=3.5mm of demo] (modes) {\textbf{(1)} find the ways (modes)};
\node[train, right=3.5mm of modes] (dp) {\textbf{(2)} label decision points};
\draw[arr] (demo) -- (modes); \draw[arr] (modes) -- (dp);
\node[infer, below=6mm of demo] (det) {at a decision point?};
\node[infer, below=6mm of modes] (prop) {probability of each mode $\pi$};
\node[infer, below=6mm of dp] (gate) {\textbf{(3)} gate: is the choice open?};
\draw[arr] (det) -- (prop); \draw[arr] (prop) -- (gate);
\draw[arr, dotted] (dp.south west) -- (det.north east); \draw[arr, dotted] (modes) -- (prop);
\node[box, fill=orange!12, below=7mm of det] (user) {user sets the mode (optional)};
\node[infer, below=7mm of prop] (tok) {\textbf{(4)} keep mode $m$ for the episode};
\node[infer, below=7mm of gate] (reg) {action chunk from state (+ mode)};
\draw[arr] (gate.south west) -- node[lab, left, pos=0.45]{yes:\\draw $m$} (tok.north east);
\draw[arr] (gate) -- node[lab, right]{no:\\no mode} (reg);
\draw[arr] (tok) -- (reg); \draw[arr, dashed] (user) -- (tok);
\node[font=\scriptsize\itshape, gray!70!black, anchor=south west] at ([yshift=0.5mm]demo.north west) {offline, from data};
\node[font=\scriptsize\itshape, blue!60!black, anchor=north west] at ([yshift=-0.5mm]user.south west) {online, during the episode (blue)};
\end{tikzpicture}
\caption{\dch{} in four ideas. Offline (gray): find the ways and the decision points in unlabeled demonstrations. Online (blue): at the first decision point a gate checks whether the choice is genuinely open; if so, one mode is drawn and kept for the rest of the episode; if not, the head acts without a mode.}
\label{fig:method}
\end{figure}

\dch{} rests on four ideas (Fig.~\ref{fig:method}).

\textbf{(1) Find the ways without labels.} Each demonstration's hand path, relative to its start, is resampled to 50 points and clustered into $M{\le}32$ modes (Gaussian mixture, size chosen by BIC). No route or order labels are used.

\textbf{(2) Learn where the choice happens.} A training state is labelled a decision point when its future action chunk has at least two modes (our re-implementation of the estimator of~\citet{mazza2026multimodality}). One small network then predicts both ``is this a decision point?'' and the probability $\pi(m\mid c_t)$ of each mode.

\textbf{(3) Choose only when the choice is genuine.} At the first decision point, a gate looks at how spread out $\pi$ is. With $\bar H{=}H(\pi)/\ln M$ the normalized entropy, versions 2--4 commit iff $\bar H{\ge}0.5$; version~5 commits iff
\begin{equation}
\bar H \;\ge\; \min\!\left(0.5,\; \ln 2/\ln M\right),
\label{eq:gate}
\end{equation}
i.e.\ also whenever the probability is spread over at least two effective modes (perplexity $e^{H(\pi)}{\ge}2$). If the state already decides the way, the gate declines and the head acts without a mode.

\textbf{(4) Keep the choice.} On commitment a mode $m{\sim}\pi$ is drawn and fed as a one-hot input to the action network at every later step of the episode. The action network is BeT-style (8 action clusters with residual offsets); in training the mode input is dropped 25\% of the time, so the network also works without a mode.

\textbf{A mode input.} Setting the mode input directly bypasses the gate and selects the way; no retraining is needed. \textbf{Cost.} On one CPU thread (Intel Core i7-13700HX, batch 1, float32), a chunk takes 0.63--0.66\,ms: 2.6--4.0$\times$ faster than our 10-step flow-matching and diffusion heads and VQ-BeT, and 1.2--1.3$\times$ slower than BeT (version-4 networks, which version~5 shares).

\section{Designing by Diagnosis: Versions 1 to 5}\label{sec:design}
\begin{figure}[t]
\centering
\begin{tikzpicture}
\begin{axis}[width=8.6cm, height=4.6cm, xmin=0.55, xmax=5.75, ymin=-0.75, ymax=0.36,
  xtick={1,2,3,4,5}, xticklabels={v1,v2,v3,v4,v5}, ylabel={held-out success $-$ BeT}, label style={font=\scriptsize},
  tick label style={font=\scriptsize}, ymajorgrids, grid style={gray!25}]
\addplot[only marks, mark=*, red!75!black, error bars/.cd, y dir=both, y explicit] coordinates {
 (1,-0.157) -= (0,0.086) += (0,0.105)
 (2,-0.111) -= (0,0.030) += (0,0.030)
 (2,-0.648) -= (0,0.041) += (0,0.042)
 (3,-0.229) -= (0,0.053) += (0,0.050)
 (4,-0.063) -= (0,0.022) += (0,0.023)};
\addplot[only marks, mark=diamond*, mark size=2.6pt, gray!70!black, error bars/.cd, y dir=both, y explicit] coordinates {
 (4,0.016) -= (0,0.039) += (0,0.040)};
\addplot[only marks, mark=*, mark size=2.6pt, blue!70!black, error bars/.cd, y dir=both, y explicit] coordinates {
 (5,0.160) -= (0,0.031) += (0,0.028)};
\draw[thick] (axis cs:0.55,0) -- (axis cs:5.75,0);
\node[font=\tiny, anchor=south west, inner sep=1pt] at (axis cs:0.57,0) {BeT};
\begin{scope}[every node/.style={font=\tiny, anchor=west, inner sep=1pt}]
\node at (axis cs:1.07,-0.157) {sorting-2};
\node at (axis cs:2.07,-0.111) {sorting-4};
\node at (axis cs:2.07,-0.648) {pushing};
\node at (axis cs:3.07,-0.229) {aligning};
\node at (axis cs:4.07,-0.075) {Kitchen};
\node at (axis cs:4.07,0.035) {stacking};
\node at (axis cs:5.07,0.160) {Kitchen};
\end{scope}
\begin{scope}[every node/.style={font=\tiny, align=center, text=gray!35!black, inner sep=1pt}]
\node at (axis cs:1.5,0.25) {$\rightarrow$ gate +\\no-mode branch};
\node at (axis cs:2.5,0.25) {$\rightarrow$ L2 loss};
\node at (axis cs:3.5,0.25) {$\rightarrow$ BeT-style\\action net};
\node at (axis cs:4.5,0.25) {$\rightarrow$ no re-select.,\\gate (\ref{eq:gate})};
\end{scope}
\end{axis}
\end{tikzpicture}
\caption{The design record: every held-out result against BeT, by version (95\% CI). Red: below BeT; gray: within uncertainty; blue: superior (registered). Gray text at top: the change each diagnosis led to in the next version (Table~\ref{tab:record}). Points of different versions are on different tasks, so this is a record, not a learning curve. Stacking: registered two-box outcome (three-box success below 0.10 for every head). Pushing was added as a second v2 test after one sorting-4 BeT cell signalled a floor risk, before any \dch{} sorting-4 result.}
\label{fig:record}
\end{figure}
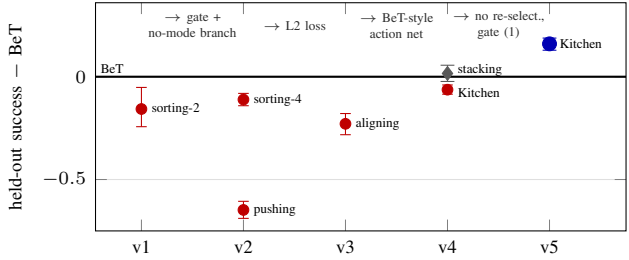

\begin{table}[t]
\centering
\caption{From diagnosis to design: what each failure revealed and what the next version changed.}
\label{tab:record}
\setlength{\tabcolsep}{2pt}\scriptsize
\begin{tabular}{@{}p{0.05\columnwidth}p{0.50\columnwidth}p{0.42\columnwidth}@{}}
\toprule
Ver. & Diagnosis (registered experiments) & Change in the next version\\
\midrule
v1 & On sorting-2 the proposal was confident, yet the head committed and timed out; re-selection, its threshold and the mode choice were ruled out as causes & v2: commit only when the proposal is uncertain; train a no-mode branch (mode dropout 0.25)\\
v2 & On pushing the regression loss limited it, not commitment (plain L1 0.10 vs.\ L2 0.68) & v3: L2 action loss\\
v3 & On aligning the gate never committed (certain proposal), so success rested on an action network that trailed BeT & v4: BeT-style action network\\
v4 & On development tasks: commitment helped on avoiding; re-selecting a mode when the arm ``stalls'' hurt (turning it off: $+0.062$ on pushing); the gate closed as clustering found more modes & v5: no stall re-selection; gate (\ref{eq:gate}); training unchanged\\
\bottomrule
\end{tabular}
\end{table}

\textbf{Why diagnose instead of guess.} A new head has many parts, and a failure rarely says which part caused it; adding parts by intuition can fix one benchmark while the real cause stays. We therefore followed a fixed loop: \emph{freeze} a version, \emph{test} it on tasks it has not seen, and, if it falls short, \emph{diagnose} with registered experiments---forcing the gate closed, forcing single modes, swapping only the loss, measuring the gate's inputs---before changing anything. This has three benefits. Each change targets a measured cause (Table~\ref{tab:record}), so the final design can be explained part by part. Held-out tests catch changes that only fit the development tasks. And failures are kept as evidence: every held-out result against BeT is in Fig.~\ref{fig:record}, including the four versions that fell below it somewhere.

\textbf{How version 5 came about.} Version~4 was registered as the last version; on held-out stacking it was comparable to BeT. Development diagnostics then found two inference-time problems (last row of Table~\ref{tab:record}), and by a disclosed departure from the plan we built version~5, which fixes only these: stall re-selection is off and the gate is (\ref{eq:gate}). Training is unchanged. Version~5 was frozen before any Kitchen result existed.

\section{Results}\label{sec:results}
\textbf{Setup.} Development tasks: the controlled task and D3IL avoiding, pushing and aligning~\citep{jia2024d3il}. Held-out for version~5: Franka Kitchen (relay kitchen, 566 demonstrations, 35 orders of the first four subtasks~\citep{gupta2019relay}) and D3IL stacking. All heads share inputs, chunk length, network size (small MLPs on the controlled task; $3{\times}1024$ MLPs or matching transformers elsewhere) and training recipe (AdamW, learning rate $10^{-4}$, batch 512, EMA 0.995, 60k steps); we evaluate the final checkpoint, never one chosen by test score. D3IL sorting, pushing, aligning and stacking use 300 test contexts per seed. Seeds: 5 (controlled), 3 (D3IL), 10 (Kitchen, 500 episodes each). Intervals are 95\% bootstraps over seeds, then episodes (10{,}000 resamples). With $\delta{=}0.032$, \emph{superior} means a gain $\ge\delta$ with the interval above 0, and \emph{non-inferior} means a lower bound above $-\delta$.

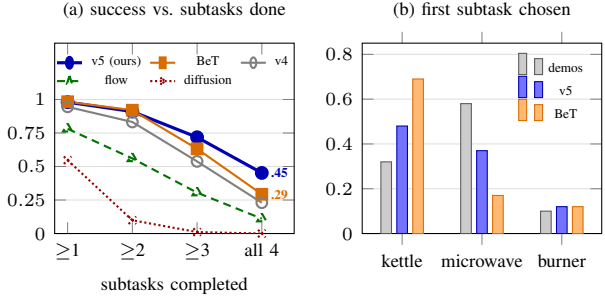
\begin{figure}[t]
\centering
\begin{tikzpicture}
\begin{groupplot}[group style={group size=2 by 1, horizontal sep=9mm},
  width=4.75cm, height=4.1cm, tick label style={font=\scriptsize}, label style={font=\scriptsize},
  title style={font=\scriptsize}, ymajorgrids, grid style={gray!25}]
\nextgroupplot[title={(a) success vs.\ subtasks done}, ymin=0, ymax=1.42, ytick={0,0.25,0.5,0.75,1}, xmin=0.8, xmax=4.5, xtick={1,2,3,4},
  xticklabels={$\ge$1,$\ge$2,$\ge$3,all 4}, xlabel={subtasks completed},
  legend style={font=\tiny, draw=none, fill=none, at={(0.5,0.99)}, anchor=north, legend columns=3, column sep=1pt, row sep=-2pt}, legend image post style={xscale=0.55}]
\addplot[mark=*, very thick, blue!70!black] coordinates {(1,0.980)(2,0.910)(3,0.718)(4,0.452)};
\addplot[mark=square*, thick, orange!85!black] coordinates {(1,0.983)(2,0.920)(3,0.632)(4,0.293)};
\addplot[mark=o, thick, gray] coordinates {(1,0.945)(2,0.832)(3,0.538)(4,0.230)};
\addplot[mark=triangle, thick, green!45!black, densely dashed] coordinates {(1,0.781)(2,0.560)(3,0.305)(4,0.111)};
\addplot[mark=x, thick, red!60!black, densely dotted] coordinates {(1,0.547)(2,0.099)(3,0.011)(4,0.000)};
\legend{v5 (ours), BeT, v4, flow, diffusion}
\node[font=\tiny\bfseries, blue!70!black, anchor=west] at (axis cs:4.0,0.452) {.45};
\node[font=\tiny\bfseries, orange!85!black, anchor=west] at (axis cs:4.0,0.293) {.29};
\nextgroupplot[title={(b) first subtask chosen}, ybar, bar width=4pt, ymin=0, ymax=0.85,
  symbolic x coords={kettle,microwave,burner}, xtick=data, enlarge x limits=0.25,
  legend style={font=\tiny, draw=none, fill=none, at={(0.98,0.98)}, anchor=north east, row sep=-2pt}]
\addplot[fill=gray!40, draw=gray!70!black] coordinates {(kettle,0.32)(microwave,0.58)(burner,0.10)};
\addplot[fill=blue!55, draw=blue!70!black] coordinates {(kettle,0.48)(microwave,0.37)(burner,0.12)};
\addplot[fill=orange!55, draw=orange!85!black] coordinates {(kettle,0.69)(microwave,0.17)(burner,0.12)};
\legend{demos, v5, BeT}
\end{groupplot}
\end{tikzpicture}
\caption{Held-out Franka Kitchen (10 seeds $\times$ 500 episodes; same network, inputs and training for every head; state: hand position and 30 robot and object coordinates). (a) Share of episodes completing at least $n$ of the four subtasks; both \dch{} versions commit in every episode. (b) Share of each first subtask: version~5 stays closer to the demonstrated mix than BeT.}
\label{fig:kitchen}
\end{figure}
\begin{table}[t]
\centering
\caption{Every head on every task (success, seed means). Shading: success relative to the best baseline in the row; bold: best baseline. BeT is the best baseline, or within noise of it, on every task, so every held-out test compares with BeT.}
\label{tab:allheads}
\setlength{\tabcolsep}{1.6pt}\scriptsize
\begin{tabular}{@{}lcccccccc|c@{}}
\toprule
Task & L1 & L2 & MIP & AR & VQ-BeT & Flow & Diff. & BeT & \dch{}\\
\midrule
avoiding & \cellcolor{orange!41}.667 & \cellcolor{orange!0}.000 & \cellcolor{orange!0}.000 & \cellcolor{orange!50}.807 & \cellcolor{orange!38}.610 & \cellcolor{orange!30}.490 & \cellcolor{orange!18}.300 & \cellcolor{orange!50}\textbf{.813} & \cellcolor{blue!12}.967$^\dagger$\\
sorting-2 & \cellcolor{orange!43}.667 & \cellcolor{orange!41}.643 & \cellcolor{orange!48}.743 & \textcolor{gray}{--} & \textcolor{gray}{--} & \textcolor{gray}{--} & \textcolor{gray}{--} & \cellcolor{orange!50}\textbf{.776} & \cellcolor{blue!12}.619 (v1)\\
sorting-4 & \cellcolor{orange!13}.043 & \cellcolor{orange!10}.034 & \cellcolor{orange!12}.041 & \cellcolor{orange!25}.083 & \textcolor{gray}{--} & \textcolor{gray}{--} & \textcolor{gray}{--} & \cellcolor{orange!50}\textbf{.166} & \cellcolor{blue!12}.054 (v2)\\
pushing & \cellcolor{orange!6}.102 & \cellcolor{orange!43}.683 & \cellcolor{orange!10}.154 & \cellcolor{orange!38}.602 & \textcolor{gray}{--} & \textcolor{gray}{--} & \textcolor{gray}{--} & \cellcolor{orange!50}\textbf{.787} & \cellcolor{blue!12}.764$^\dagger$\\
aligning & \cellcolor{orange!0}.001 & \cellcolor{orange!20}.134 & \cellcolor{orange!3}.020 & \cellcolor{orange!36}.246 & \cellcolor{orange!19}.131 & \cellcolor{orange!36}.247 & \cellcolor{orange!5}.032 & \cellcolor{orange!50}\textbf{.342} & \cellcolor{blue!12}.402$^\dagger$\\
stacking$^a$ & \cellcolor{orange!50}\textbf{.140} & \cellcolor{orange!6}.017 & \cellcolor{orange!0}.000 & \cellcolor{orange!38}.107 & \cellcolor{orange!46}.130 & \cellcolor{orange!6}.016 & \cellcolor{orange!0}.000 & \cellcolor{orange!50}.139 & \cellcolor{blue!12}.154 (v4)\\
Kitchen $p_4$ & \textcolor{gray}{--} & \textcolor{gray}{--} & \textcolor{gray}{--} & \textcolor{gray}{--} & \textcolor{gray}{--} & \cellcolor{orange!19}.111 & \cellcolor{orange!0}.000 & \cellcolor{orange!50}\textbf{.293} & \cellcolor{blue!12}\textbf{.452} (v5)\\
\midrule
ms/chunk$^b$ & \textcolor{gray}{--} & \textcolor{gray}{--} & \textcolor{gray}{--} & \textcolor{gray}{--} & 2.19 & 2.12 & 2.60 & 0.50 & \cellcolor{blue!12}0.66 (v4)\\
\bottomrule
\end{tabular}\\[2pt]
\raggedright\scriptsize $^a$Two-box outcome (three-box success below 0.10 for every head). \dch{}: version~5 where it ran ($^\dagger$ development), else the version tested held-out (Fig.~\ref{fig:record}). Sorting-2: contexts 0--299 (AR, VQ-BeT and generative heads were run on a 30-context subset only). $^b$Median per chunk on stacking, one CPU thread (version~5 shares version~4's networks). --: not run or not timed.
\end{table}
\textbf{Which baseline is strongest, and is it fair?} We ran eight baseline heads (Table~\ref{tab:allheads}). BeT had the highest success among them on every task, or was within noise of the highest (avoiding: AR 0.807 vs.\ 0.813; stacking: L1 0.140 vs.\ 0.139); BeT is the registered reference, and a registered rule would have replaced it only if another head exceeded it; none did. Our BeT on Kitchen (0.293) lies in its registered range (published 0.44; the released BeT run ends at 0.10 and peaks at 0.34). Our diffusion and flow heads fall far below the published Diffusion Policy~\citep{chi2023diffusion}, so the registered calibration check failed, although the released checkpoint reaches $p_4{=}1.00$ on our installation (Section~\ref{sec:lessons}, finding~3). We therefore claim only the controlled comparison with BeT.

\textbf{Q1. Does \dch{} exceed BeT on a task it never saw?} On Kitchen, yes (Fig.~\ref{fig:kitchen}a). Version~5 completes all four subtasks in 45.2\% of episodes against 29.3\% for BeT: $+0.160$ \ci{+0.129}{+0.188}, superior by a registered test; it is also far above flow matching (0.111) and diffusion (0.000), the other heads run on Kitchen (descriptive). Version~4 fell below BeT ($-0.063$ \ci{-0.085}{-0.040}); version~5 minus version~4 is $+0.222$ \ci{+0.197}{+0.248} (descriptive). Both commit in every episode and share checkpoints, so the gain comes from the inference changes. Post hoc, we trace it to stall re-selection, which fired about four times per version-4 episode because Kitchen manipulation moves the arm little in the plane. Version~5's first subtasks are closer to the demonstrated mix than BeT's (Fig.~\ref{fig:kitchen}b).

\textbf{Q2. Does the mode input set the behavior?} On Kitchen, when a mode is forced (episodes 0--99, seeds 0--2), the first completed subtask is that mode's main subtask in 95.4\% of successful episodes; forced single modes range from 0.07 to 0.84 in $p_4$ (descriptive). BeT has no comparable input. On the controlled task, forcing works under large start variation (5 of 5 seeds) but not when the start fixes the route (purity 0.58, 0 of 5 seeds) or with 100 demonstrations (descriptive).

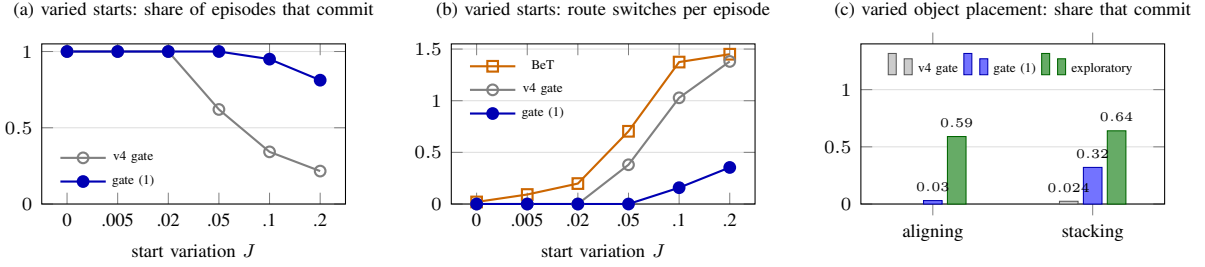
\begin{figure*}[t]
\centering
\begin{tikzpicture}
\begin{groupplot}[group style={group size=3 by 1, horizontal sep=14mm},
  width=5.6cm, height=3.7cm, tick label style={font=\scriptsize}, label style={font=\scriptsize},
  title style={font=\scriptsize}, legend style={font=\tiny, draw=none, fill=none}, ymajorgrids, grid style={gray!25}]
\nextgroupplot[title={(a) varied starts: share of episodes that commit}, ymin=0, ymax=1.05, legend pos=south west,
  xtick={1,...,6}, xticklabels={0,.005,.02,.05,.1,.2}, xlabel={start variation $J$}]
\addplot[mark=o, thick, gray] coordinates {(1,1)(2,1)(3,1)(4,0.620)(5,0.342)(6,0.216)};
\addplot[mark=*, thick, blue!70!black] coordinates {(1,1)(2,1)(3,1)(4,1)(5,0.950)(6,0.812)};
\legend{v4 gate, gate (\ref{eq:gate})}
\nextgroupplot[title={(b) varied starts: route switches per episode}, ymin=0, ymax=1.55, legend pos=north west,
  xtick={1,...,6}, xticklabels={0,.005,.02,.05,.1,.2}, xlabel={start variation $J$}]
\addplot[mark=square, thick, orange!80!black] coordinates {(1,0.020)(2,0.092)(3,0.198)(4,0.704)(5,1.374)(6,1.450)};
\addplot[mark=o, thick, gray] coordinates {(1,0)(2,0)(3,0)(4,0.380)(5,1.028)(6,1.380)};
\addplot[mark=*, thick, blue!70!black] coordinates {(1,0)(2,0)(3,0)(4,0)(5,0.158)(6,0.354)};
\legend{BeT, v4 gate, gate (\ref{eq:gate})}
\nextgroupplot[title={(c) varied object placement: share that commit}, ybar, bar width=7pt, ymin=0,
  ymax=1.4, ytick={0,0.5,1}, symbolic x coords={aligning,stacking}, xtick={aligning,stacking}, enlarge x limits=0.45,
  legend style={at={(0.5,0.98)}, anchor=north, legend columns=3}, nodes near coords, every node near coord/.append style={font=\tiny, /pgf/number format/fixed, /pgf/number format/precision=3}]
\addplot[fill=gray!40, draw=gray!70!black] coordinates {(stacking,0.024)};
\addplot[fill=blue!55, draw=blue!70!black] coordinates {(aligning,0.03)(stacking,0.32)};
\addplot[fill=green!45!black!60, draw=green!40!black] coordinates {(aligning,0.59)(stacking,0.64)};
\legend{v4 gate, gate (\ref{eq:gate}), exploratory}
\end{groupplot}
\end{tikzpicture}
\caption{Where commitment breaks and how far diagnosis repaired it. (a,b) Varied starts, controlled task (two ways, 400 demonstrations, 5 seeds; development, gate change adopted on these data). (c) Varied object placement, a step toward cluttered scenes; success did not change detectably, though non-inferiority is not shown. ``v4 gate'': threshold 0.5 of versions 2--4.}
\label{fig:dose}
\end{figure*}
\textbf{Q3. Where does commitment break, and can it be fixed?} Diagnosis found three weak spots: varied starts (repaired), varied object placement (partly repaired) and few demonstrations (open: with 100 instead of 400 demonstrations, commitment nearly vanished, 0.02).

\emph{(a) Varied start positions: found and repaired} (controlled task, development). Version~4's commitment fell as start positions varied (Fig.~\ref{fig:dose}a: 1.00 to 0.22). A diagnosis registered before any fix found why: the clustering split the same two routes into more modes ($M$ from 12 to 30--31), so dividing the entropy by $\ln M$ pushed it below 0.5, although the network still split evenly between the two \emph{routes} (route entropy 0.82--0.90). Gate (\ref{eq:gate}), adopted by four criteria fixed before its runs, restores commitment to 0.81 at the largest variation, cuts route switches under per-step re-planning from 1.38 to 0.35 per episode (Fig.~\ref{fig:dose}b) and raises success there from 0.744 to 0.912 (BeT 0.740); on the development tasks it stays within $\delta$. These results are in-sample, on one task.

\emph{(b) Varied object placement, a step toward cluttered scenes: partly repaired.} When each demonstration has its own object layout (aligning, stacking), the network was nearly certain at the decision point (median $\bar H$ 0.003 on aligning), although the scene barely predicts inside versus outside pushing (5-nearest-neighbor accuracy 0.521, majority 0.500); so the gate rarely commits (Fig.~\ref{fig:dose}c; stacking: 2.4\% of episodes with the version-4 gate). Gate (\ref{eq:gate}) raises stacking commitment to 0.32 (target 0.5 not met), with no detectable harm on the floor-limited three-box outcome ($+0.006$ \ci{-0.004}{+0.016}; a weak test). An exploratory variant that trains the proposal on the mode mix of the 10 nearest demonstrations goes further: commitment 0.03 to 0.59 on aligning and 0.32 to 0.64 on stacking, with success changes of $-0.002$ \ci{-0.063}{+0.060} and $-0.014$ \ci{-0.036}{+0.007}. Its no-harm tests were registered on point estimates; by the interval rule neither is shown non-inferior.

\textbf{Q4. What about the other tasks?} Version~5 minus BeT on development tasks: 0.000 on the controlled task (both 1.000); $+0.153$ \ci{+0.097}{+0.217} on avoiding, a commitment effect (forcing version~4's gate closed costs $-0.163$); $+0.060$ \ci{+0.016}{+0.104} on aligning, not a commitment effect (the gate committed in 3\% of episodes); and $-0.022$ \ci{-0.076}{+0.029} on pushing. On held-out stacking, three-box success stays below 0.10 for every head (best 0.043); on the registered two-box outcome, version~4 was comparable to BeT ($+0.016$ \ci{-0.023}{+0.056}) and VQ-BeT and above L2, but its gate committed in only 2.4\% of episodes. \dch{} (version~4) did not exceed BeT on stacking; version~5 was not compared with BeT there by a registered test.

\section{Findings Useful Beyond \dch{}}\label{sec:lessons}

\textbf{1. Multimodality measures did not predict when the head matters} (negative result). Forward predictions from measured multimodality missed on robomimic (six of six datasets), avoiding and sorting (cross-task slope $-0.375$, descriptive); benchmarks need tasks where the state gives no hint which way to take.

\textbf{2. How often a policy re-plans can matter as much as which head it uses.} On avoiding, L2 and MIP regression went from 0.00 to 1.00 when re-planning every 2 instead of every 10 steps (descriptive). Head comparisons should report the execution horizon.

\textbf{3. Head rankings depend on the action representation and protocol} (descriptive). With one MLP backbone, relative actions, one observation step and the final checkpoint, our diffusion and flow heads reached $p_4$ 0.000 and 0.111 on Kitchen, and other sampling settings kept them at or below 0.12; the released Diffusion Policy checkpoint (absolute joint targets, two observation steps, a U-Net, 5000 epochs, checkpoint selection by test score) reached 1.00 on the same installation,; its authors report that diffusion policies depend on position control.

\textbf{4. Normalizing uncertainty by the number of modes is a trap.} An entropy gate divided by $\ln M$ closes as clustering finds more modes, even when the real choice is unchanged; testing the effective number of options (perplexity $\ge2$) reduced this on the controlled task, though not to the registered level on stacking. This applies to any method that thresholds normalized entropy.

\textbf{5. Progress heuristics must match the task} (post hoc). A stall rule based on planar arm motion, harmless on navigation-like tasks, coincided with a 0.22 loss in four-subtask success on Kitchen. \textbf{6. A confident proposal can hide an open choice} (post hoc): under object-placement variation the proposal was near-certain although the scene barely predicts the choice.

\textbf{7. Commitment is not recovery.} With the chosen route blocked by an unseen wall (controlled task), every sampling head, \dch{} included, succeeds in 0.33--0.48 of episodes; recovery needs re-deciding, which commitment does not provide. \textbf{8. Commitment is cheap:} 0.63--0.66\,ms per chunk, near BeT (Table~\ref{tab:allheads}).

\section{Scope, Departures and \Later}\label{sec:limits}
\textbf{Scope.} All evidence is simulated, and our diffusion and flow heads are weak on Kitchen in our setting. The start-variation fix is shown on one controlled task; commitment under object-placement variation is only partly restored, and fully only by an exploratory variant; few demonstrations remain unsolved; and we make no claim about cluttered household scenes.
\textbf{Departures from the registered plan.} Version~4 was registered as the last version and a later entry kept it as the primary result; version~5 was then explored by lifting a registered gate (no key factor had been found), candidates were added after the candidate list was closed, and a second gate variant was evaluated outside its trigger and not adopted; two other candidates were not selected; Kitchen baselines and version~4 were trained on Kitchen before version~5 was frozen, though no Kitchen result existed then; the Kitchen superiority test was registered after the original claim plan; a version-5 action network retrained with mode dropout 0.5 fell below version~5 on Kitchen ($-0.073$ \ci{-0.103}{-0.043}); a registered ceiling test for re-deciding at subtask boundaries passed against BeT (oracle 0.427) but was not above version~5, and that version was not built by author decision; the real-robot test was postponed, discarding demonstrations recorded before the postponement; and the robot uses one camera instead of two because two could not stream together.
\textbf{Real robot (planned).} An SO-ARM\,101 moves a cube past an obstacle with one or two demonstrated sides (100 demonstrations each), seen by one fixed external camera ($320{\times}240$) plus joint positions. Predictions fixed in advance for version~5 against BeT at two sides, 50 rollouts each: success above BeT by $\delta$; higher success when the chosen side is blocked; commitment in at least 90\% of rollouts; fewer mid-route switches when re-planning every 10 and every 2 steps; and, with 20 forced rollouts per side, at least 90\% of successful forced rollouts on the commanded side (lower bound 0.80). With 50 rollouts per arm, only large success differences (about 0.26) are detectable. If the gate does not commit on hardware, no commitment claim is made. Control-loop latency on the robot is not yet measured.

\section{Conclusion}
The action head matters when demonstrations contain several ways and the state does not say which to take. \dch{} then selects a way once, at a learned decision point, commits when its proposal is uncertain, and exposes the choice as a mode input. Designed by diagnosing each version's failures, its fifth version exceeded BeT, our strongest baseline, on held-out Kitchen; its fourth did not on stacking; and diagnosis located, and partly repaired, where commitment breaks.

\bibliographystyle{plainnat}
\bibliography{references}

\ifappendix
\appendix
This appendix appears only in the preprint. It lists every registered test used in the paper with its outcome, the departures from the registered plan, implementation details, and additional tables. C-S denotes the controlled 2-D task of Section~\ref{sec:when}, and ``task'' replaces the word ``carrier'' used in our logs. All intervals are 95\% two-level bootstraps (10{,}000 resamples); $\delta{=}0.032$.

\subsection{Registered tests and outcomes}
Table~\ref{tab:reg} lists every registered test used in the paper, with its value and verdict as recorded before any later change.
\begin{table*}[h]
\centering\scriptsize
\setlength{\tabcolsep}{3pt}
\resizebox{\textwidth}{!}{%
\begin{tabular}{@{}lllll@{}}
\toprule
Id & Task & Test & Value \ci{lo}{hi} & Verdict\\
\midrule
P1 & C-S & slope of $\Delta$ on $\log_2 k$ & $+0.333$ \ci{+0.134}{+0.437} & supported\\
P2 & C-S $k{=}1$ & $|\Delta|\le\delta$ & $-0.032$ \ci{-0.039}{-0.025} & inconclusive\\
P5 & C-S & $\Delta(k{=}2)-\Delta(k{=}2\mathrm{F})\ge\delta$ & $+0.768$ \ci{+0.363}{+0.981} & supported\\
HM-1 & C-S & $\Delta$ falls with start jitter (0, 0.005, 0.02) & $+0.724 > +0.126 > -0.056$ & supported\\
HM-3 & C-S & $\Delta$(open loop) $>\Delta$(closed loop) $+\delta$, jitter 0.02 & $+0.143$ \ci{+0.097}{+0.189} & supported\\
HM-4 & C-S & blocked route: deterministic $\le 0.1$, noise heads $\ge 0.3$ & 0.00 vs.\ 0.42 & supported\\
HM-5 & avoiding & $\Delta(e{=}10)-\Delta(e{=}2)\ge\delta$ & $+0.319$ \ci{-0.040}{+0.967} & inconclusive\\
OP & robomimic, avoiding, sorting-2 & forward-chained prediction of $\Delta$ & missed on every task & OP-3\\
NP-1b & sorting-2 (vision) & L2($e{=}2$) $-$ L2($e{=}10$) $\ge\delta$ & $+0.244$ \ci{+0.089}{+0.400} & supported\\
\midrule
EH-1x & sorting-2 & v1 $-$ BeT, non-inferiority (contexts 0--299) & $-0.157$ \ci{-0.243}{-0.052} & not supported\\
V2-S1 / V2-S4 & sorting-4 & v2 $-$ L1 / v2 $-$ BeT, non-inferiority & $+0.011$ / $-0.111$ & supported / not supported\\
V2-S3 & sorting-4 & share of v2 episodes without commitment $\ge 0.70$ & 0.994 & supported\\
V2-P1 / V2-P2 & pushing & v2 $-$ L1 superiority / v2 $-$ BeT non-inferiority & $+0.037$ / $-0.648$ & supported / not supported\\
V2-P3 & pushing & v2 committed share $\ge 0.90$ & 1.000 & supported\\
V3-A2 / V3-A3 & aligning & v3 $-$ v2 superiority / v3 $-$ BeT non-inferiority & $+0.112$ / $-0.229$ & supported / not supported\\
V3-A5 & aligning & v3 commits among decision-point entries $\ge 0.90$ & 0.000 & not supported\\
V4-S1-2box & stacking & v4 $-$ BeT, non-inferiority & $+0.016$ \ci{-0.023}{+0.056} & supported\\
V4-S2-2box / V4-S4-2box & stacking & v4 $-$ L2 / v4 $-$ v3, superiority & $+0.138$ / $+0.124$ & supported\\
X-S1/2/3-2box & stacking & v4 $-$ flow / diffusion / VQ-BeT, non-inferiority & $+0.139$ / $+0.154$ / $+0.024$ & supported\\
\midrule
Dose & C-S & commitment falls with start jitter / with start-determined route & 0.784 drop / $-0.136$ & pattern B (jitter only)\\
G (a)--(d) & C-S, development & gate (\ref{eq:gate}) adoption criteria & all four met & adopted into v5\\
S-1 / S-2 & stacking & v5 commit $\ge 0.5$ / v5 $-$ v5-C $\ge-\delta$ & 0.319 / $+0.006$ \ci{-0.004}{+0.016} & not met / met\\
K-1 & Kitchen & v5 $-$ BeT on $p_4$ & $+0.160$ \ci{+0.129}{+0.188} & superior\\
K-2 & Kitchen & v4 $-$ BeT on $p_4$ & $-0.063$ \ci{-0.085}{-0.040} & inconclusive or inferior\\
K-3 & Kitchen & v5 $-$ v4 (descriptive) & $+0.222$ \ci{+0.197}{+0.248} & --\\
K-4 & Kitchen & committed share $\ge 0.5$ & v5 1.000, v4 1.000 & met\\
K-5 & Kitchen & forced-mode purity of the first subtask (descriptive) & 0.954, 0.956, 0.953 & --\\
Calib. & Kitchen & our diffusion $p_4\ge 0.80$, our BeT in $[0.25, 0.65]$ & 0.000; 0.293 & not met\\
X-1 & Kitchen / stacking & v5-X $-$ v5 lower bound $>-\delta$ & $-0.073$ \ci{-0.103}{-0.043} / $-0.006$ \ci{-0.016}{+0.004} & not met / met\\
X-2 & Kitchen & v5-X standing against BeT not below v5's & both superior & met\\
O-1--O-4 & aligning, stacking & v5-S commit $\ge 0.5$; v5-S $-$ v5 $\ge-\delta$ (point) & 0.592, 0.638; $-0.002$, $-0.014$ & all met\\
C-1 & Kitchen & oracle re-decision at subtasks $-$ BeT & $+0.134$ \ci{+0.109}{+0.159} & pass (v6 not built)\\
\bottomrule
\end{tabular}}
\caption{Registered tests used in the paper. v5-X: version 5 with a regressor retrained with token dropout 0.5. v5-S: proposal trained on the mode mix of the 10 nearest demonstrations. Oracle: version 5 re-deciding the mode whenever the environment reports a completed subtask.}
\label{tab:reg}
\end{table*}

\begin{table}[!t]
\centering\scriptsize
\setlength{\tabcolsep}{3pt}
\begin{tabular}{@{}lccc@{}}
\toprule
Task (role for v5) & v5 & BeT & v5 $-$ BeT\\
\midrule
C-S, $J{=}0$ (development) & 1.000 & 1.000 & $+0.000$\\
avoiding (development) & 0.967 & 0.813 & $+0.153$ \ci{+0.097}{+0.217}\\
pushing (development) & 0.764 & 0.787 & $-0.022$ \ci{-0.076}{+0.029}\\
aligning (development) & 0.402 & 0.342 & $+0.060$ \ci{+0.016}{+0.104}\\
stacking, 3 boxes (held-out) & 0.044 & 0.043 & --\\
Kitchen $p_4$ (held-out) & 0.452 & 0.293 & $+0.160$ \ci{+0.129}{+0.188}\\
\bottomrule
\end{tabular}
\caption{Version 5 against BeT on every task it was run on.}
\end{table}

\begin{table}[!t]
\centering\scriptsize
\setlength{\tabcolsep}{2.5pt}
\begin{tabular}{@{}lcccccc@{}}
\toprule
C-S dataset & commit v4 & commit v5 & succ.\ v4 & succ.\ v5 & BeT & steer (seeds)\\
\midrule
$J{=}0$ & 1.000 & 1.000 & 1.000 & 1.000 & 1.000 & 1.000 (5/5)\\
$J{=}0.005$ & 1.000 & 1.000 & 1.000 & 1.000 & 1.000 & 1.000 (5/5)\\
$J{=}0.02$ & 1.000 & 1.000 & 1.000 & 1.000 & 1.000 & 0.989 (5/5)\\
$J{=}0.05$ & 0.620 & 1.000 & 1.000 & 1.000 & 1.000 & 0.953 (5/5)\\
$J{=}0.1$ & 0.342 & 0.950 & 1.000 & 1.000 & 1.000 & 0.931 (5/5)\\
$J{=}0.2$ & 0.216 & 0.812 & 0.744 & 0.912 & 0.740 & 0.921 (5/5)\\
$J{=}0.1$, $q{=}0.75$ & 0.316 & 0.832 & 0.998 & 1.000 & 1.000 & 0.891 (1/5)\\
$J{=}0.1$, $q{=}1$ & 0.478 & 0.782 & 1.000 & 1.000 & 1.000 & 0.579 (0/5)\\
$J{=}0.1$, 100 demos & 0.022 & 0.574 & 0.996 & 0.998 & 1.000 & 0.826 (0/5)\\
\bottomrule
\end{tabular}
\caption{C-S dose study at $e{=}10$ (5 seeds). Steering columns refer to version 4 (forcing bypasses the gate, so version 5 steers identically).}
\end{table}

\begin{table}[!t]
\centering\scriptsize
\begin{tabular}{@{}lcc@{}}
\toprule
Head (one CPU thread, batch 1) & stacking & aligning\\
\midrule
BeT & 0.495 & 0.524\\
\dch{} (v4 networks) & 0.659 & 0.631\\
VQ-BeT & 2.185 & 1.615\\
Flow matching (10 steps) & 2.119 & 1.892\\
Diffusion (10 DDIM steps) & 2.604 & 2.331\\
\bottomrule
\end{tabular}
\caption{Median time per chunk (ms), Intel Core i7-13700HX, float32; one sample call, excluding observation reading and environment steps.}
\end{table}

\subsection{Departures from the registered plan}
\begin{itemize}
\item Version~5 exists by a deviation: the registered rule allowed a version~5 only if a diagnostic found a key factor; none was found, and the author lifted the rule. Candidates were chosen from development diagnostics, selected by a registered rule (stall re-selection off), and extended by an inference-only gate change after a registered diagnosis of start variation; the gate change was adopted by four criteria fixed before its runs. Version~5 was frozen before any Franka Kitchen evaluation result existed (Kitchen baselines and version~4 were already being trained) and before any real-robot data was used.
\item Version~4 had been registered as the last version, and a later entry kept it as the primary result; the Kitchen superiority test for version~5 was registered after the original claim plan.
\item A second gate variant (test at the first call) was evaluated after its trigger condition was not met; it failed the adoption criteria and was not adopted.
\item The exploratory arms v5-X and v5-S were registered after version~5 was frozen; they do not replace it.
\item After seeing Kitchen results, a diagnosis of version~4's Kitchen deficit and a ceiling test for a sixth version were registered; the sixth version was dropped because the ceiling was not above version~5.
\item The calibration check failed; its investigation (sampling settings, the released checkpoint) was registered stage by stage.
\item The real-robot test was postponed; demonstrations recorded before the postponement are not used.
\item On the real robot, one external camera replaces two because of a hardware limit found before any registered demonstration.
\end{itemize}

\subsection{Implementation details}
\emph{Modes.} Paths of the end-effector (first two state dimensions, standardized) relative to their start, resampled to 50 points; PCA to $\min(8,n-1)$ dimensions; spherical Gaussian mixture, $K\in[2,\min(32,\lfloor n/3\rfloor)]$ by BIC, 5 initializations. \emph{Labels.} The conditional-mode estimator (horizon 10, bandwidth 0.5, PCA to 2, BIC mixture, cluster mass $\ge 0.025$) on every 5th training state on C-S and avoiding, every 10th (at most 5{,}000) elsewhere; a state is a decision point if it has at least two modes. \emph{Networks.} Proposal and detector: one MLP, 3 layers of 256 units; regressor: BeT-style head with hidden width 1024 on input $[c_t, f, z]$. AdamW, learning rate $10^{-4}$, weight decay $10^{-4}$, batch 512, EMA 0.995, 60k steps per stage. \emph{Kitchen.} State: end-effector position and the 30 robot and object coordinates of one observation (goal coordinates dropped); action: the recorded 9-dimensional relative command; 280 steps, robot noise ratio 0.1, episodes 0--499 per seed.

\subsection{Additional results}
\begin{figure}[!t]
\centering
\begin{tikzpicture}
\begin{axis}[xbar=0pt, width=7.6cm, height=4.2cm, xmin=-0.15, xmax=0.27,
  symbolic y coords={push,align,avoid,cs,kit}, ytick={push,align,avoid,cs,kit},
  yticklabels={{pushing (dev.)},{aligning (dev.)},{avoiding (dev.)},{controlled (dev.)},{Kitchen (held-out)}},
  tick label style={font=\scriptsize}, label style={font=\scriptsize}, xlabel={version 5 $-$ BeT (success)},
  xmajorgrids, grid style={gray!25}, enlarge y limits=0.15]
\addplot[only marks, mark=*, blue!70!black, error bars/.cd, x dir=both, x explicit] coordinates {
 (0.160,kit) -= (0.031,0) += (0.028,0)
 (0.000,cs)
 (0.153,avoid) -= (0.056,0) += (0.064,0)
 (0.060,align) -= (0.044,0) += (0.044,0)
 (-0.022,push) -= (0.054,0) += (0.051,0)};
\draw[thick] (axis cs:0,push) -- (axis cs:0,kit);
\draw[dashed, gray] (axis cs:0.032,push) -- (axis cs:0.032,kit);
\draw[dashed, gray] (axis cs:-0.032,push) -- (axis cs:-0.032,kit);
\end{axis}
\end{tikzpicture}
\caption{Version~5 against BeT on every task where both ran (95\% CI; dashed: $\pm\delta$). Development tasks were used for design, so only Kitchen is a held-out test. Stacking is omitted because version~5 was not compared with BeT there by a registered test.}
\label{fig:v5all}
\end{figure}
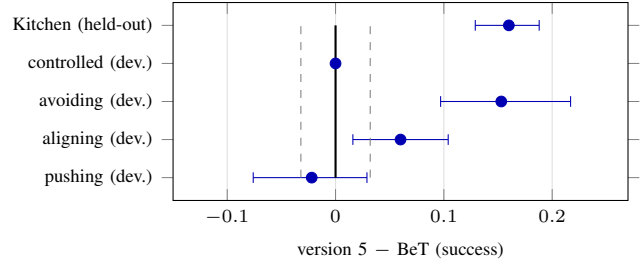
Fig.~\ref{fig:v5all} shows version~5 against BeT on every task where both ran.

\emph{Kitchen, further arms (descriptive, 10 seeds).} Version~4 with the gate held closed: $p_4$ 0.248. Version~5 re-deciding at every completed subtask (oracle trigger): 0.427. v5-X: 0.379. The released Diffusion Policy checkpoint under its own evaluation on our installation: $p_4$ 1.00 (50 test episodes). The released BeT run: $p_4$ 0.10 at its last evaluation, 0.168 averaged over its last five, 0.34 at its best.

\fi

\end{document}